\documentclass[letterpaper]{article}

\usepackage{aaai2027}

\usepackage[hyphens]{url}
\usepackage{graphicx}
\usepackage{natbib}
\usepackage{caption}

\usepackage{algorithm}
\usepackage{algorithmic}

\usepackage{newfloat}
\usepackage{listings}

\DeclareCaptionStyle{ruled}{
    labelfont=normalfont,
    labelsep=colon,
    strut=off
}

\floatstyle{ruled}
\newfloat{listing}{tb}{lst}{}
\floatname{listing}{Listing}

\usepackage{booktabs}

\usepackage{amsmath}
\usepackage[dvipsnames]{xcolor}
\usepackage{subcaption}
\usepackage{multirow}
\usepackage{colortbl}
\usepackage{array}

\title{
WildShadowRemover: In-the-Wild Video Shadow Removal\\
via Detail-Preserving Video Diffusion Models
}

\author{
Jiamin Xu\textsuperscript{\rm 1},
Cong Wang\textsuperscript{\rm 1},
Zheng Dong\textsuperscript{\rm 2},
Chi Wang\textsuperscript{\rm 3},
Renshu Gu\textsuperscript{\rm 1},
Weiwei Xu\textsuperscript{\rm 3},
Gang Xu\textsuperscript{\rm 1}
}

\affiliations{
\textsuperscript{\rm 1}
Hangzhou Dianzi University, Hangzhou, China\\
\textsuperscript{\rm 2}
Zhejiang Sci-Tech University, Hangzhou, China\\
\textsuperscript{\rm 3}
State Key Laboratory of CAD\&CG, Zhejiang University, Hangzhou, China
}

\begin{document}

\maketitle

\begin{abstract}

Video shadow removal in the wild remains challenging due to complex illumination, diverse shadow appearances, and limited training data. Despite its importance to numerous vision and graphics applications, it remains largely unexplored in unconstrained real-world scenarios. To address this gap, we present WildShadowRemover, a framework that adapts a pretrained video diffusion model for robust video shadow removal via LoRA fine-tuning. To preserve fine image details while retaining the model's powerful generative prior, we augment the frozen VAE decoder with a detail injection module and introduce a shadow-mask-guided frequency-decomposed modulation module to selectively restore high-frequency textures while suppressing shadow artifacts. Monocular depth priors from Depth Anything 3 further provide geometry-aware guidance under challenging lighting conditions. We also construct WildShadow, a large-scale paired video shadow removal dataset and benchmark, covering diverse synthetic scenes. Extensive experiments demonstrate that our method outperforms existing approaches in shadow removal quality and temporal consistency, producing temporally coherent shadow-free videos with superior visual quality and strong generalization across challenging in-the-wild scenarios.

\end{abstract}

\section{Introduction}
\label{sec:intro}

Shadows are a common illumination phenomenon caused by the occlusion of light transport. While they provide valuable cues for scene understanding~\cite{zhang1999shape}, they also adversely affect a wide range of vision and graphics applications, including object tracking~\cite{sanin2010improved}, intrinsic decomposition~\cite{li2018learning, li2022physically, nestmeyer2020learning, ye2023intrinsicnerf}, and image editing~\cite{zhang2021noshadow}. Despite recent progress, shadow removal in in-the-wild environments, such as urban streets, natural landscapes, and UAV-captured scenes, remains challenging due to complex illumination, out-of-view occlusions, and highly diverse shadow appearances.

\begin{figure}[t]
\centering
\includegraphics[width=1.0\linewidth]{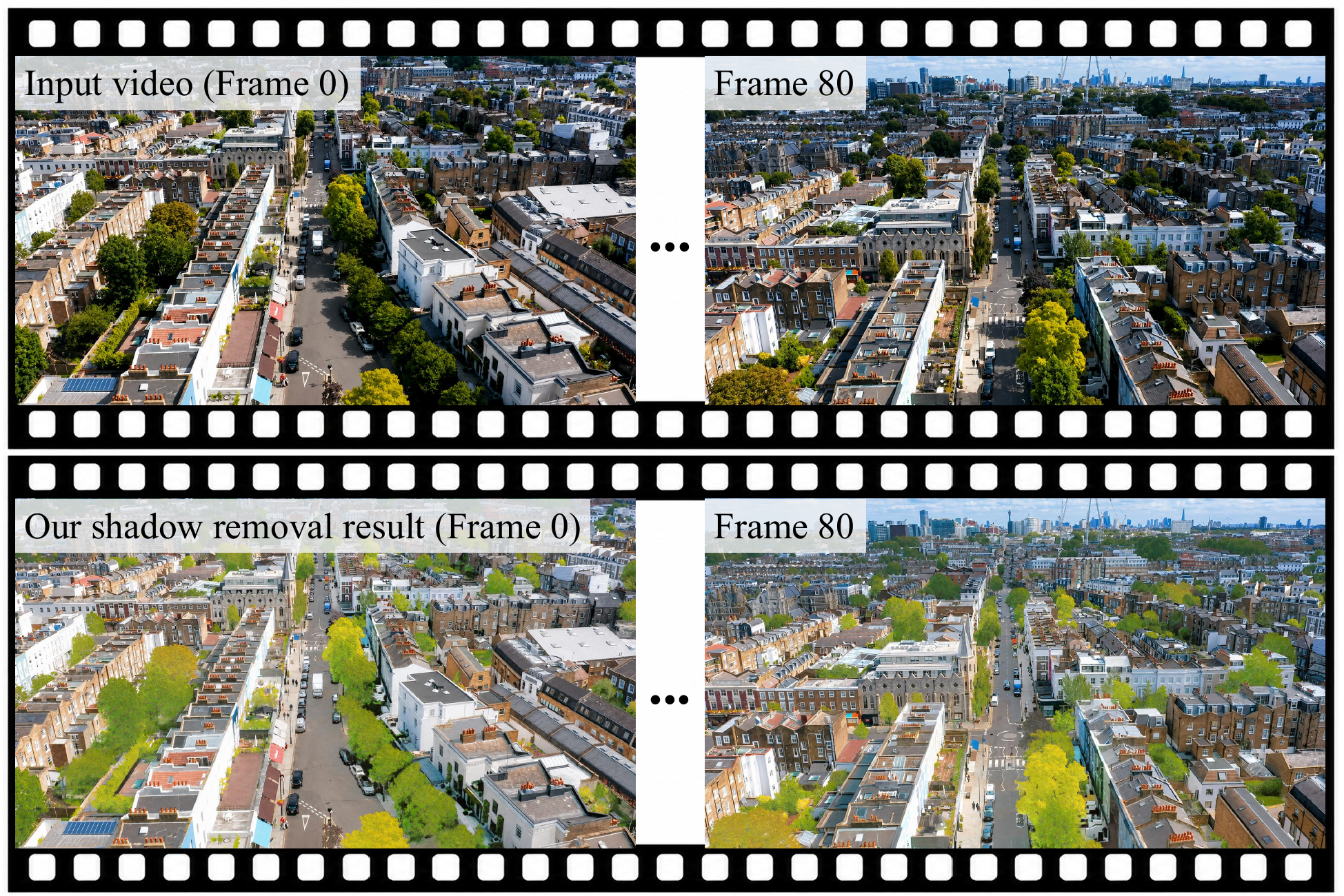}
\caption{\textbf{Our proposed WildShadowRemover effectively removes complex shadows from in-the-wild videos.} Given a shadowed input video (top), our method produces shadow-free results (bottom) while preserving scene details and maintaining temporal consistency across frames.}
\label{fig:teaser}
\end{figure}

We argue that limited generalization to  in-the-wild environments is the primary obstacle to robust in-the-wild shadow removal, arising from two key factors: training data and foundation models. On the data side, paired shadow/shadow-free datasets remain scarce and lack sufficient scene diversity. Early datasets, such as ISTD and SRD, are collected under controlled conditions. More recent datasets, such as OmniSR~\cite{xu2024omnisr} and Infinigen Indoors~\cite{infinigen2024indoors}, generate paired training data through physically based rendering, but primarily focus on indoor environments. As a result, current datasets still fail to cover the diversity and complexity of real-world scenes, leazding to poor generalization to challenging in-the-wild scenarios, such as urban streets and UAV-captured videos.

On the foundation model side, recent methods, such as StableShadowRemoval(StableSR)~\cite{xu2025detail}, have begun to exploit image diffusion models pretrained on Internet-scale data. In contrast, recent video diffusion models (VDMs)~\cite{ho2022video,wan2025wan} learn richer spatiotemporal priors and have demonstrated remarkable performance in video depth estimation~\cite{xu2025diffusion} and video editing~\cite{yu2025objectmover,xiao2026relit}. We hypothesize that training on large-scale videos enables VDMs to implicitly capture the intrinsic correlation between objects and their shadows. Consequently, pretrained VDMs naturally preserve temporally coherent shadow evolution consistent with object motion while maintaining coherent scene content, making them a promising foundation model for robust shadow identification and removal. Furthermore, these spatiotemporal priors facilitate the generation of temporally consistent shadow-free videos, benefiting downstream applications such as 3D content creation.

Motivated by these observations, we build our framework upon a pretrained VDM and adapt it to the video shadow removal task through parameter-efficient LoRA fine-tuning~\cite{hu2022lora}. Rather than directly generating shadow-free frames, our goal is to effectively exploit the powerful generative prior of the pretrained VDM while preserving fine image details. To this end, we introduce a detail injection module, inspired by~\cite{xu2025detail}, to restore high-frequency details without compromising the learned generative prior. To prevent shadow artifacts from being reintroduced during detail injection, we further propose a shadow-mask-guided frequency-decomposed modulation module that selectively modulates high- and low-frequency features according to the predicted shadow mask, preserving shadow-free textures while suppressing shadow-related details. Furthermore, we leverage geometric priors from Depth Anything 3 (DA3)~\cite{lin2025depth} to provide geometry-aware guidance, enabling more reliable shadow removal under complex illumination.

To support the proposed framework, we construct a large-scale synthetic video shadow removal dataset, termed \emph{WildShadow}. The indoor subset is generated by rendering video clips from 3D-FRONT scenes, following a pipeline similar to OmniSR~\cite{xu2024omnisr}. For outdoor scenes, we employ the iCity~\cite{icity2024} Blender plugin together with Infinigen~\cite{raistrick2023infinite} to synthesize diverse urban and natural environments. To comprehensively evaluate generalization, we additionally build a benchmark consisting of Habitat Synthetic Scenes (HSSD)~\cite{khanna2023hssd}, and newly constructed iCity scenes~\cite{icity2024}.


Overall, our contributions can be summarized as follows.
\begin{itemize}

\item We present WildShadowRemover, a diffusion-based framework for robust in-the-wild video shadow removal that combines detail injection, shadow-mask-guided frequency modulation, and geometry-aware guidance to preserve fine textures while removing complex shadows.

\item We introduce WildShadow, a large-scale synthetic dataset for video shadow removal, together with a comprehensive benchmark covering diverse synthetic scenes.

\item Extensive experiments demonstrate the effectiveness of our method on both synthetic and real-world benchmarks, achieving improved generalization, temporal consistency, and visual quality compared with existing approaches.

\end{itemize}

\section{Related Work}
\label{sec:related}

\subsection{Image Shadow Removal}

Deep learning has significantly advanced image shadow removal by learning mappings from shadow images to shadow-free images. Early deep learning methods establish CNN-based shadow removal frameworks and improve shadow detection and restoration~\cite{qu2017deshadownet,hu2019mask,cun2020towards,fu2021auto,zhu2022bijective}. Recently, transformer-based methods, such as ShadowFormer~\cite{guo2023shadowformer}, HomoFormer~\cite{xiao2024homoformer}, and DMTN~\cite{liu2023decoupled}, further enhance shadow removal by exploiting long-range dependencies. Other approaches incorporate physical priors, frequency-domain modeling, and geometry-semantic guidance to improve robustness under complex illumination, including ShadowRefiner~\cite{dong2024shadowrefiner}, OmniSR~\cite{xu2024omnisr}, DenseSR~\cite{densesr2025}, and PhaSR~\cite{phaser2026}.

Recently, diffusion models have shown strong potential for shadow removal by leveraging powerful generative priors. Existing diffusion-based methods, including ShadowDiffusion~\cite{guo2023shadowdiffusion}, DeS3~\cite{jin2024des3}, Diff-Shadow~\cite{luo2024diff}, and latent diffusion-based approaches~\cite{mei2024latent}, formulate shadow removal as a conditional generation problem. StableShadowRemoval~\cite{xu2025detail} further adapts a pretrained Stable Diffusion model with latent-space fine-tuning and detail-preserving decoder enhancement, improving texture restoration and real-world generalization. However, most existing methods focus on single-image shadow removal and cannot explicitly model temporal consistency across video frames.

\subsection{Video Shadow Removal}

Video shadow removal remains largely under-explored due to the challenges of temporal consistency, dynamic object motion, and varying illumination conditions. Early approaches rely on temporal illumination modeling, such as spatio-temporal illumination transfer methods~\cite{zhang2017video}, but often struggle with complex real-world dynamic scenes.

Recent methods formulate video shadow removal as a restoration problem by exploiting spatial and temporal correlations. Chen et al.~\cite{chen2024pstnet} proposed PSTNet, which jointly models physical, spatial, and temporal characteristics of video shadows, and introduced the SVSRD-85 synthetic paired video shadow removal dataset. However, SVSRD-85 is limited in scale and its source data are not publicly available, making reproduction and large-scale training challenging. Moreover, existing benchmarks mainly focus on synthetic scenarios and lack sufficient diversity for unconstrained real-world environments.

Recent advances in large-scale video diffusion models~\cite{ho2022video,wan2025wan} have demonstrated strong capabilities in modeling spatiotemporal dynamics and visual priors. These models learn rich temporal and appearance priors from large-scale video collections, providing a promising direction for robust video shadow removal. In this work, we adapt a pretrained video diffusion model with geometry-aware guidance and shadow-aware detail restoration to achieve temporally consistent and high-fidelity shadow-free video generation in unconstrained environments.


\begin{figure*}[t]
    \centering
    \includegraphics[
        width=0.95\textwidth
    ]{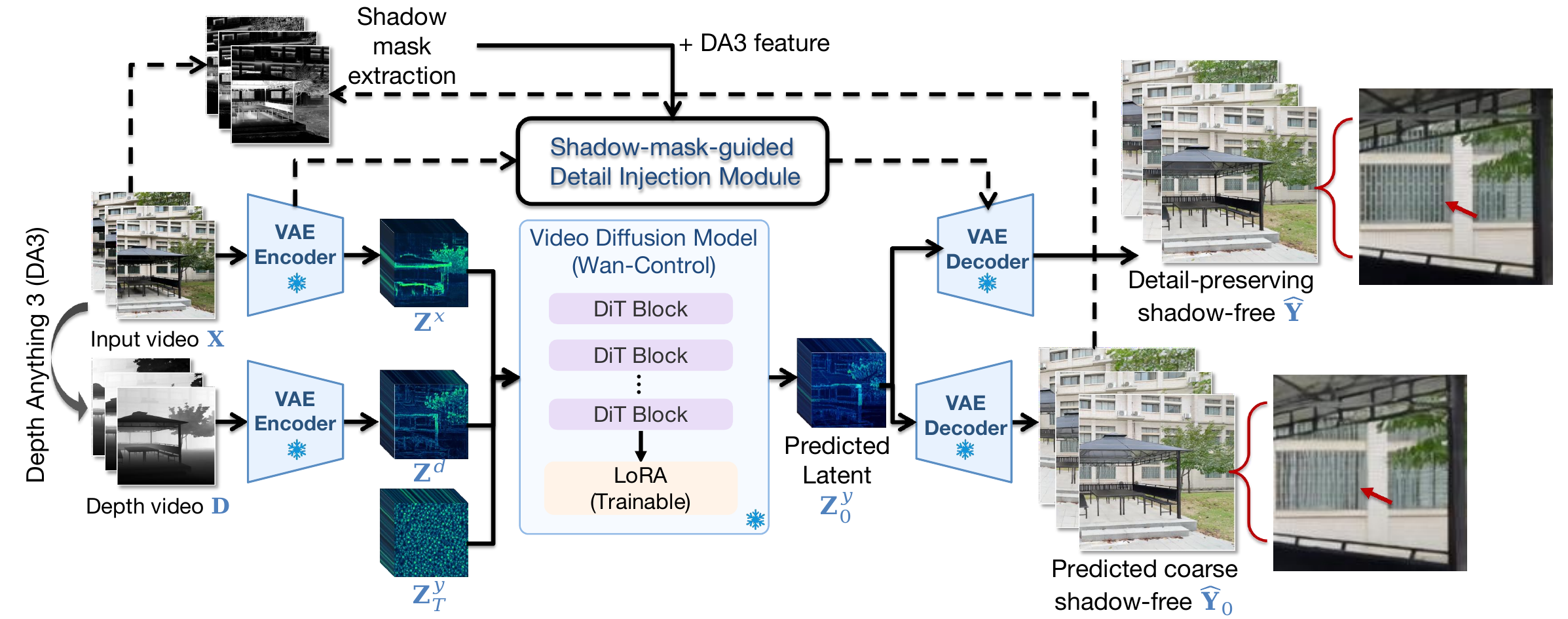}
    \caption{\textbf{Overview of the proposed video shadow removal framework.} A latent video diffusion model predicts a coarse shadow-free result, followed by a shadow-mask-guided detail injection module for detail restoration. Red arrows highlight enlarged regions showing the improved detail preservation of the final output.}
    \label{fig:pipeline}
\end{figure*}

\section{Proposed Method}
\label{sec:method}

\subsection{Overview}
\label{sec:overview}

We denote the input shadow video and its paired shadow-free target as
$\mathbf{X}=\{\mathbf{x}^{t}\}_{t=1}^{T}$ and
$\mathbf{Y}=\{\mathbf{y}^{t}\}_{t=1}^{T}$,
respectively, where $T$ denotes the number of frames. Our framework performs shadow removal in the latent space of a pretrained Wan-Control video diffusion model. Specifically, the diffusion transformer is conditioned on the latent representations of the input shadow video $\mathbf{X}$ and the corresponding DA3 depth maps $\mathbf{D}$. We adapt the transformer through LoRA fine-tuning~\cite{hu2022lora} to predict a coarse shadow-free latent representation $\hat{\mathbf{Z}}^{y}_{0}$, which is subsequently decoded by the VAE decoder into the corresponding coarse shadow-free video $\hat{\mathbf{Y}}_{0}$.

To recover fine-grained details, we keep the pretrained VAE frozen and augment its decoder with a shadow-mask-guided detail injection module. However, directly injecting appearance features may reintroduce residual shadow artifacts due to the entanglement of shadows and image details in the feature space. To address this issue, we introduce frequency-decomposed modulation (FDM), which separates appearance features into low- and high-frequency components and selectively modulates them using a shadow mask estimated from the input shadow video $\mathbf{X}$ and the coarse shadow-free prediction $\hat{\mathbf{Y}}_{0}$. This enables detail restoration while suppressing shadow artifacts (Fig.~\ref{fig:pipeline}).

\subsection{Latent Video Diffusion for Shadow Removal}
\label{sec:latent_video_generation}

We build our framework upon Wan~\cite{wan2025wan} and adopt its Wan-Control variant, a latent video diffusion model based on the flow-matching paradigm that supports controllable video generation using RGB-video and depth-video conditions. The model consists of a spatiotemporal variational autoencoder (VAE), a diffusion transformer (DiT), and a T5 text encoder. The VAE maps videos between the pixel and latent spaces, while the DiT learns a conditional velocity field in the latent space. Specifically, we adapt the pretrained Wan-Control model from RGB-and-depth-controlled video generation to conditional shadow-to-shadow-free video translation by conditioning the diffusion model on the input shadow video and its estimated depth video. All pretrained model components remain frozen except for a small set of trainable LoRA~\cite{hu2022lora} parameters inserted into the DiT, enabling parameter-efficient adaptation.

\paragraph{Geometry-Conditioned Latent Encoding.}
Let $\mathcal{E}$ and $\mathcal{D}$ denote the pretrained video VAE encoder and decoder, respectively. Given an input shadow video $\mathbf{X}$ and the shadow-free video $\mathbf{Y}$, we obtain their video latent representations:
\begin{equation}
    \mathbf{Z}^{x}
    =
    \mathcal{E}(\mathbf{X}),
    \qquad
    \mathbf{Z}^{y}
    =
    \mathcal{E}(\mathbf{Y}).
    \label{eq:video_latent_encoding}
\end{equation}
Since shadow formation is closely related to scene geometry, we estimate a monocular depth video from the input shadow video using Depth Anything 3 (DA3)~\cite{lin2025depth}:
\begin{equation}
    \mathbf{D}
    =
    \mathrm{DA3}(\mathbf{X}).
    \label{eq:depth_estimation}
\end{equation}
We replicate the normalized depth maps across three channels to form an RGB-like video and encode it with the pretrained video VAE, yielding the depth-video latent $\mathbf{Z}^{d}=\mathcal{E}(\mathbf{D})$. All three latents, $\mathbf{Z}^{x}$, $\mathbf{Z}^{y}$, and $\mathbf{Z}^{d}$, have 16 channels and share the same spatiotemporal resolution of $\frac{W}{8}\times\frac{H}{8}\times\left(1+\frac{T-1}{4}\right)$. Here, $T=4N+1$ is required by the temporal downsampling structure of the VAE. Further details of the VAE architecture can be found in Wan~\cite{wan2025wan}.

\paragraph{Flow-Matching Fine-Tuning.}
We adapt the pretrained DiT using latent-space flow matching. Given the target shadow-free video latent $\mathbf{Z}^{y}$, a Gaussian noise latent
$\boldsymbol{\epsilon}\sim\mathcal{N}(\mathbf{0},\mathbf{I})$,
and a continuous timestep
$t\sim\mathcal{U}(0,1)$,
we construct a linear interpolation between the noise distribution and the target latent distribution:
\begin{equation}
    \mathbf{Z}_{t}
    =
    (1-t)\boldsymbol{\epsilon}
    +
    t\mathbf{Z}^{y}.
    \label{eq:interpolated_video_latent}
\end{equation}
The corresponding target velocity along this interpolation path is:
\begin{equation}
    \mathbf{V}_{t}
    =
    \frac{\partial \mathbf{Z}_{t}}{\partial t}
    =
    \mathbf{Z}^{y}
    -
    \boldsymbol{\epsilon}.
    \label{eq:target_video_velocity}
\end{equation}
Here, $\mathbf{Z}_{t}$ denotes an interpolated latent video rather than a conventionally noise-corrupted latent used in DDPM-based formulations.

Let $f_{\theta}$ denote the DiT velocity predictor. At each timestep, the interpolated target latent is concatenated channel-wise with the fixed shadow-video and depth-video conditions:
\begin{equation}
    \hat{\mathbf{V}}_{t}
    =
    f_{\theta}
    \left(
        \operatorname{Concat}
        \left[
            \mathbf{Z}_{t},
            \mathbf{Z}^{x},
            \mathbf{Z}^{d}
        \right],
        \mathbf{c},
        t
    \right),
    \label{eq:video_velocity_prediction}
\end{equation}
where $\mathbf{c}$ denotes the text embedding of the fixed prompt ``remove shadows from the video''.


The model is optimized using the standard flow-matching objective:
\begin{equation}
\mathcal{L}_{\mathrm{FM}}
=
{E}_{t,\boldsymbol{\epsilon}}
\left[
\left\|
\hat{\mathbf{V}}_{t}
-
\mathbf{V}_{t}
\right\|_2^2
\right].
\label{eq:flow_matching_loss}
\end{equation}

\subsection{Shadow-mask-guided Detail Injection Module}
\label{sec:detail_decoding}

Although the pretrained video diffusion model effectively removes shadows, the decoded videos often suffer from the loss of fine image details, such as textures and thin structures, as shown in Fig.~\ref{fig:pipeline}. Directly injecting encoder features from the input shadow video can restore these details, but also risks reintroducing shadow patterns. To address this issue, we estimate a coarse shadow mask to guide shadow-aware detail restoration.

Specifically, the latent prediction from the diffusion transformer is first decoded by the frozen pretrained VAE decoder to obtain a coarse shadow-free video $\hat{\mathbf{Y}}_0$. We then estimate a soft shadow mask by comparing the input shadow video $\mathbf{X}$ with the coarse prediction $\hat{\mathbf{Y}}_0$ as $\mathbf{M}=1-\mathbf{X}/(\hat{\mathbf{Y}}_0+\epsilon)$, and resize it to the resolution of each decoder stage to obtain $\mathbf{m}_i$. This mask provides spatial guidance for selectively restoring appearance details during decoding.

To enhance detail restoration, we augment the frozen Wan VAE decoder with a detail injection module. At the $i$-th decoder stage, let $\mathbf{d}_i$, $\mathbf{e}_{n-i}$, and $\mathbf{a}_i$ denote the decoder feature, the corresponding encoder feature, and the DA3 feature, respectively. We first concatenate the decoder and encoder features to obtain $\mathbf{f}_i=[\mathbf{d}_i,\mathbf{e}_{n-i}]$, which is then processed by the proposed Frequency-Decomposed Modulation (FDM) module under the guidance of the shadow mask $\mathbf{m}_i$ and the DA3 feature $\mathbf{a}_i$.

\paragraph{Frequency-Decomposed Modulation.} Given an input feature $\mathbf{x}$, FDM first decomposes it into low-frequency and high-frequency components using a fixed channel-wise $5\times5$ mean filter:
\begin{equation}
\mathbf{x}^{L}
=
\mathbf{x}*\mathbf{k},
\qquad
\mathbf{x}^{H}
=
\mathbf{x}-\mathbf{x}^{L},
\qquad
\mathbf{k}
=
\frac{1}{25}\mathbf{1}_{5\times5},
\end{equation}
where $*$ denotes channel-wise convolution.

The resized shadow mask $\mathbf{m}_i$ is then fed into a lightweight modulation network $g(\cdot)$ to predict frequency-specific affine parameters:
\begin{equation}
(\gamma^{L},\beta^{L},\gamma^{H},\beta^{H})
=
g(\mathbf{m}_i),
\end{equation}
which independently modulate the low- and high-frequency components as:
\begin{equation}
\mathcal{F}_{\mathrm{FDM}}(\mathbf{x},\mathbf{m}_i)
=
(\gamma^{L}\odot\mathbf{x}^{L}+\beta^{L})
+
(\gamma^{H}\odot\mathbf{x}^{H}+\beta^{H}),
\end{equation}
where $\odot$ denotes element-wise multiplication.

\paragraph{Detail Injection Process.}
We apply the proposed detail injection process at each decoder stage
$i\in[1,n-1]$. Specifically, the decoder feature $\mathbf{d}_i$ and
its corresponding encoder feature $\mathbf{e}_{n-i}$ are first
concatenated and processed by FDM:
\begin{equation}
{\mathbf h}_i
=
\mathcal{F}_{\mathrm{FDM}}
\left(
[\mathbf d_i,\mathbf e_{n-i}],
\mathbf m_i
\right).
\end{equation}
To provide geometry-aware guidance under complex illumination, the corresponding DA3 feature is first projected to a compact feature representation using a $1\times1$ convolution and then concatenated with the modulated feature. The fused feature is subsequently processed by an RRDB~\cite{wang2018esrgan} to aggregate appearance and geometry information:
\begin{equation}
\mathbf r_i
=
\mathrm{RRDB}
\left(
[
{\mathbf h}_i,
{\mathbf a}_i
]
\right),
\end{equation}
where ${\mathbf a}_i$ denotes the projected DA3 feature.

Finally, the predicted residual is further refined using another FDM before being injected into the frozen decoder through a residual connection:
\begin{equation}
\widetilde{\mathbf d}_i
=
\mathbf d_i
+
\mathbf w_i
\odot
\mathcal F_{\mathrm{FDM}}
\left(
\mathbf r_i,
\mathbf m_i
\right),
\end{equation}
where $\mathbf{w}_i$ is a learnable channel-wise gate initialized to zero. This design enables the decoder to selectively recover high-frequency image details while effectively suppressing the reintroduction of shadow artifacts.

\subsection{Our Video Dataset}
\label{sec:dataset}

\begin{figure}[t]
    \centering    \includegraphics[width=1.0\linewidth]{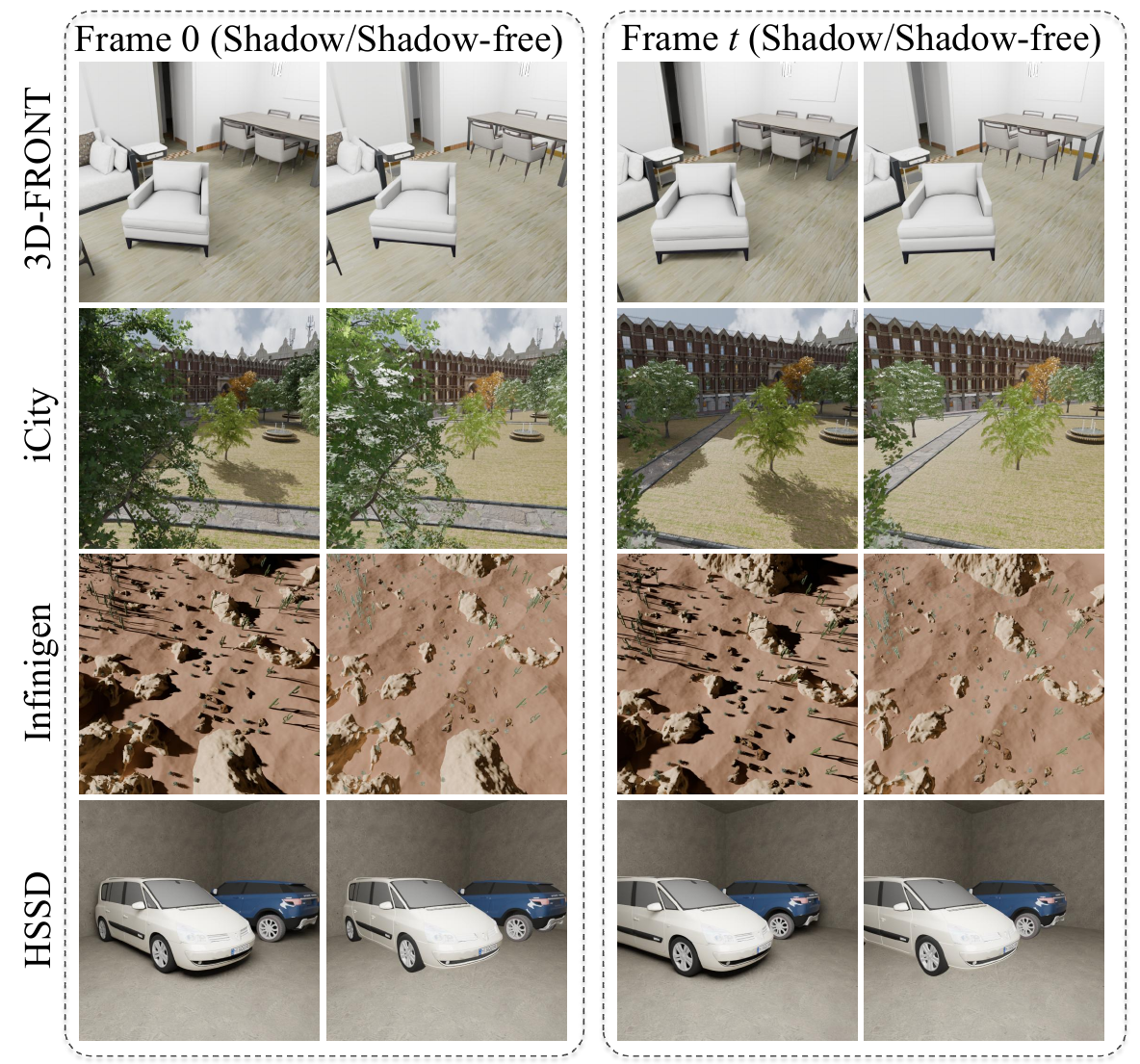}
    \captionof{figure}{\textbf{Our video dataset.} Each row shows two representative frames from paired shadowed input and shadow-free ground-truth videos.}
\label{fig:dataset_overview}
\end{figure}

We construct \emph{WildShadow}, a large-scale synthetic dataset for video shadow removal (Fig.~\ref{fig:dataset_overview}). The dataset covers diverse indoor, urban outdoor, and natural outdoor environments. Following the rendering protocol of OmniSR~\cite{xu2024omnisr}, indoor scenes are rendered from 3D-FRONT~\cite{fu20213dfront} and HSSD~\cite{khanna2023hssd}. Urban and natural outdoor scenes are generated using the iCity Blender plugin and Infinigen~\cite{raistrick2023infinite}, respectively. For outdoor scenes, shadow-free targets are generated by removing only shadows cast by direct illumination. All video clips contain 21--60 consecutive frames with frame-wise aligned shadow/shadow-free pairs.

The training set consists of 4,500 video clips from 3D-FRONT, 900 clips from iCity, and 700 clips from Infinigen. The test set contains 400, 100, and 100 clips from the corresponding datasets, respectively. In addition, we include 50 challenging indoor video clips rendered from HSSD~\cite{khanna2023hssd} exclusively for evaluation. 


\subsection{Implementation  Details}
\label{sec:training_inference}

To facilitate training, following~\cite{xu2025diffusion}, we adopt a two-stage training strategy. Our model is trained on four NVIDIA RTX 4090 GPUs using PyTorch 2.5.1 and CUDA 12.1, with DeepSpeed and distributed data parallelism. We use the AdamW optimizer with $\beta_1=0.9$, $\beta_2=0.999$, a weight decay of $0.01$, and $\epsilon=10^{-8}$. A constant learning rate of $1\times10^{-4}$ without warmup is used in both stages. In the first stage, we fine-tune the Wan-Control DiT for latent shadow-free generation while keeping the Wan VAE and text encoder frozen. Only LoRA adapters inserted into the attention projection and FFN layers are optimized. The LoRA rank and scaling factor are both set to 64. The per-GPU batch size is 1, resulting in a global batch size of 4. 

In the second stage, we freeze the first-stage DiT and Wan VAE backbone and optimize only the detail injection modules, which contain approximately 42.94M trainable parameters. The DA3 features are injected into the first two modules. The effective global batch size is 8, using a per-GPU batch size of 1 and gradient accumulation over 2 steps. Further details are provided in the supplementary material.

\begin{table*}[t]
\centering
\small
\caption{\textbf{Quantitative comparison on public benchmarks and WildShadow-I.} Best and second-best results are highlighted in bold and underline, respectively.}
\label{tab:quantitative_comparison}
\small
\begin{tabular}{l c cc cc cc cc}
\toprule
\multirow{2}{*}{Method} & \multirow{2}{*}{Mask-free}
& \multicolumn{2}{c}{ISTD+}
& \multicolumn{2}{c}{SRD}
& \multicolumn{2}{c}{INS}
& \multicolumn{2}{c}{WildShadow-I} \\
\cmidrule(lr){3-4}
\cmidrule(lr){5-6}
\cmidrule(lr){7-8}
\cmidrule(lr){9-10}
& 
& PSNR$\uparrow$ & SSIM$\uparrow$
& PSNR$\uparrow$ & SSIM$\uparrow$
& PSNR$\uparrow$ & SSIM$\uparrow$
& PSNR$\uparrow$ & SSIM$\uparrow$ \\
\midrule
ShadowFormer~\cite{guo2023shadowformer}
 & No
& \underline{35.46} & 0.971
& 32.90 & 0.958
& 28.62 & 0.963
& -- & -- \\

DMTN~\cite{liu2023decoupled}
 & No
& 32.23 & 0.966
& 33.77 & 0.968
& 28.83 & 0.969
& -- & -- \\

ShadowDiffusion~\cite{guo2023shadowdiffusion}
 & No
& 34.63 & 0.967
& \textbf{34.73} & \underline{0.970}
& 29.12 & 0.966
& -- & -- \\

HomoFormer~\cite{xiao2024homoformer}
 & No
& \textbf{35.72} & \textbf{0.977}
& \underline{34.36} & \textbf{0.977}
& 28.98 & 0.965
& -- & -- \\

\midrule

Refusion~\cite{luo2023refusion}
 & Yes
& 32.41 & 0.961
& 31.60 & 0.949
& 28.13 & 0.958
& -- & -- \\

DeS3~\cite{jin2024des3}
 & Yes
& 31.39 & 0.957
& 34.11 & 0.968
& 27.89 & 0.947
& 24.57 & 0.899 \\

OmniSR~\cite{xu2024omnisr}
 & Yes
& 33.34 & 0.970
& 32.87 & 0.969
& 30.38 & 0.973
& \underline{28.46} & \underline{0.945} \\

StableSR~\cite{xu2025detail}
 & Yes
& 35.19 & \underline{0.974}
& 33.63 & 0.968
& \underline{30.56} & \underline{0.975}
& 27.66 & 0.938 \\

PhaSR~\cite{phaser2026}
 & Yes
& 34.48 & 0.960
& -- & --
& 30.38 & 0.961
& 27.10 & 0.922 \\
\midrule
\textbf{Ours}
 & Yes
& 34.72 & 0.971
& 31.72 & 0.961
& \textbf{31.06} & \textbf{0.977}
& \textbf{29.28} & \textbf{0.961} \\
\bottomrule
\end{tabular}
\end{table*}

\begin{figure*}[t]
    \centering
    \includegraphics[width=0.85\linewidth]{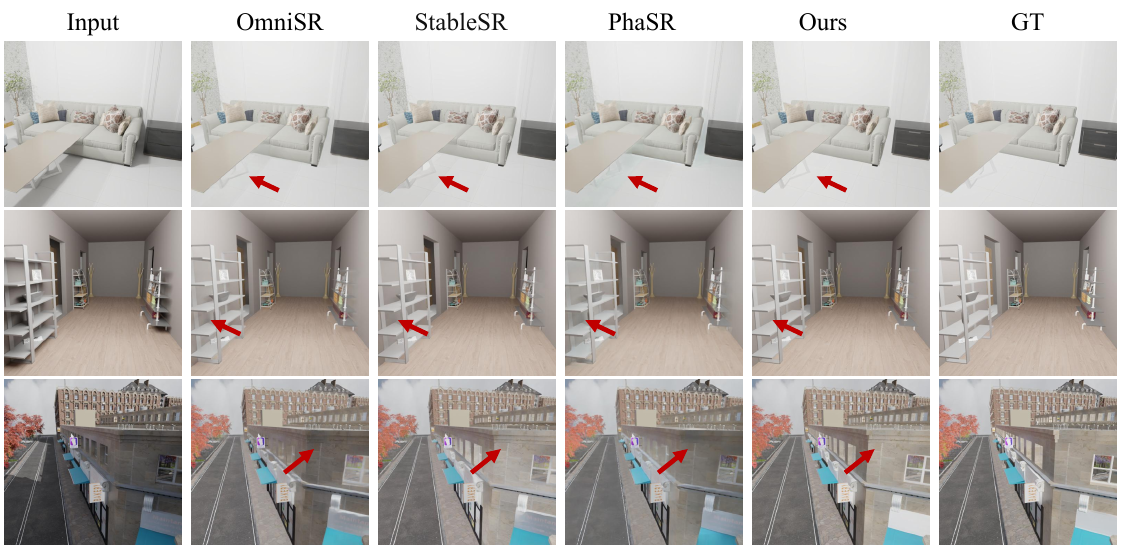}
    \caption{\textbf{Qualitative comparison on INS, HSSD, and iCity.} For this comparison, all methods are trained on our WildShadow-I training set. Our method achieves more effective shadow removal while preserving color consistency and fine-grained scene details. Red arrows indicate regions with clearer shadow removal.}
    \label{fig:sota_qualitative}
\end{figure*}

\section{Experiments}
\label{sec:experiments}

\subsection{Experimental Setup}
\label{sec:experimental_setup}

Since existing shadow-removal methods are predominantly designed for single images, we first evaluate our approach in the conventional image-based setting. Although designed for video shadow removal, our approach can be directly applied to single images without architectural modification, as the underlying Wan model supports single-frame inputs. We conduct experiments on three established benchmarks, ISTD+~\cite{wang2018stacked,le2019shadow}, SRD~\cite{qu2017deshadownet}, and INS~\cite{xu2024omnisr}, enabling direct comparison with existing methods.

To further assess generalization across diverse scene domains, we construct WildShadow-I, a unified image dataset consisting of 36,000 paired shadow/shadow-free training images from INS, iCity, and Infinigen, together with a multi-domain test set covering indoor, urban outdoor, and natural scenes, including HSSD. Further details are provided in the supplementary material. We additionally qualitatively evaluate our method on real-world images from DL3DV~\cite{ling2023dl3dv} and self-captured UAV videos.

\paragraph{Metrics and Baseline Configuration.}
Following prior work~\cite{fu2021auto,le2020shadow,guo2023shadowformer}, quantitative image evaluation is conducted at a resolution of $256\times256$. We report Peak Signal-to-Noise Ratio (PSNR) and Structural Similarity (SSIM)~\cite{wang2004image}, using the MATLAB evaluation code adopted by Zhu et al.~\cite{zhu2022efficient}. For public benchmarks, we follow the original training and evaluation protocols of each method and report the results from their corresponding papers. For our proposed in-the-wild dataset, all methods are retrained and evaluated using the same data splits and experimental protocols to ensure a fair comparison. Each baseline is implemented based on its official codebase with the training configurations recommended by the original authors.

\subsection{Comparisons}
\label{sec:sota_comparison}

We compare our method with representative mask-based and mask-free shadow-removal approaches. The mask-based methods include ShadowFormer~\cite{guo2023shadowformer}, DMTN~\cite{liu2023decoupled}, ShadowDiffusion~\cite{guo2023shadowdiffusion}, and HomoFormer~\cite{xiao2024homoformer}. The mask-free methods include Refusion~\cite{luo2023refusion}, DeS3~\cite{jin2024des3}, OmniSR~\cite{xu2024omnisr}, StableSR~\cite{xu2025detail}, and PhaSR~\cite{phaser2026}. Since our method does not require an input shadow mask, we focus the main discussion on comparisons with mask-free methods.

\paragraph{Results on public and our benchmarks.}
Table~\ref{tab:quantitative_comparison} shows that our method achieves competitive performance on conventional image shadow-removal benchmarks. While image-specific methods obtain slightly better results on ISTD+ and SRD, our method achieves the best performance on INS, which contains complex indoor shadows under both direct and indirect illumination. It obtains 31.06~dB PSNR and 0.977 SSIM, demonstrating the effectiveness of latent diffusion priors, geometric guidance, and detail restoration for handling complex illumination variations.

On our proposed WildShadow-I dataset, our method achieves the best performance with 29.28~dB PSNR and 0.961 SSIM, outperforming all reproduced baselines. Compared with the second-best OmniSR, our method improves PSNR by 0.82 dB and SSIM by 0.016. Since all methods are trained and evaluated under the same protocol, these improvements demonstrate the effectiveness of our framework for shadow removal across diverse scene domains.

Figure~\ref{fig:sota_qualitative} presents qualitative comparisons on INS, HSSD, and iCity. Existing methods often leave residual shadows or introduce brightness inconsistencies, especially in regions with soft illumination changes and complex occlusions. In contrast, our method produces cleaner shadow-free results while better preserving color consistency, object boundaries, and fine-grained scene details across diverse indoor and outdoor scenes.

\begin{figure}[t]
    \centering
    \includegraphics[width=\columnwidth]{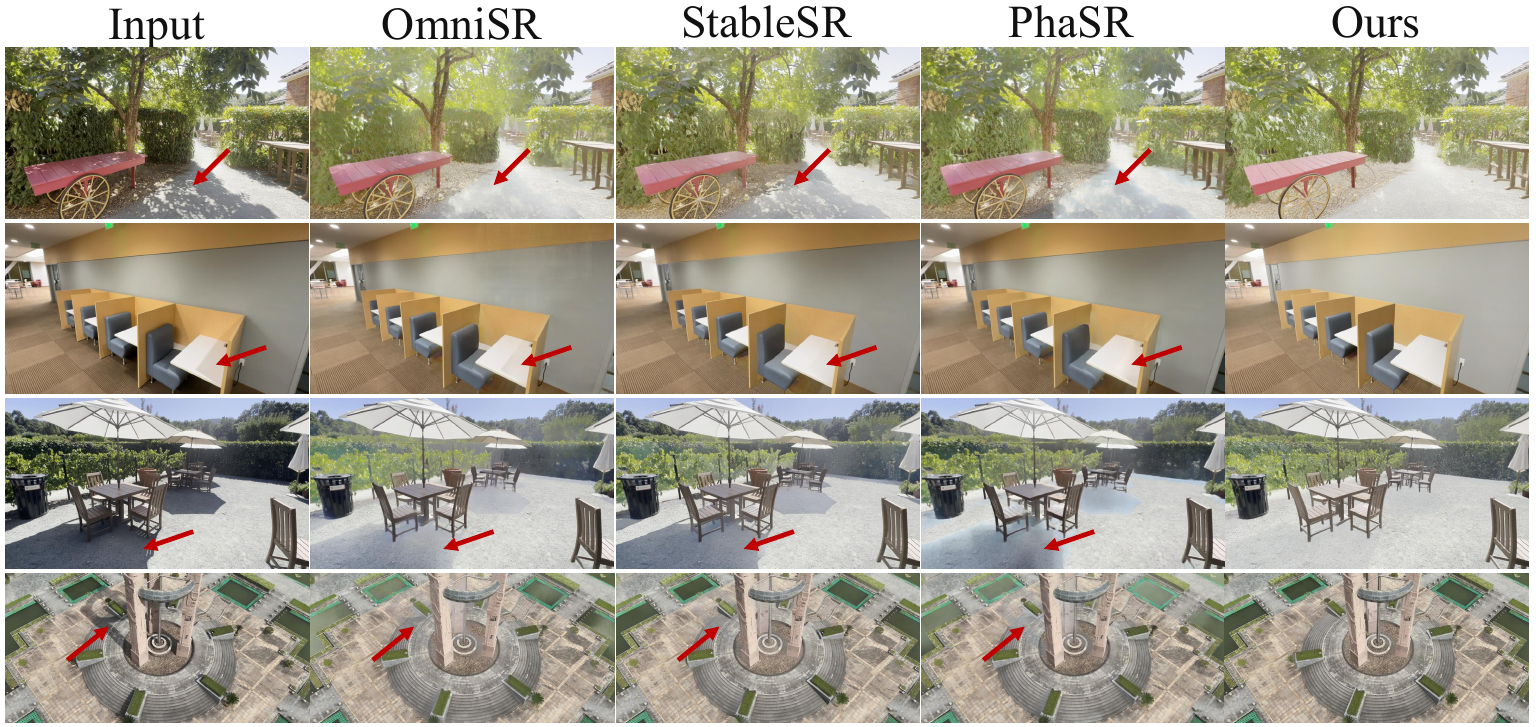}
    \caption{Qualitative comparison on real-world DL3DV videos and our captured UAV videos without ground-truth references.}
    \label{fig:dl3dv_compare}
\end{figure}

\begin{table}[t]
\centering
\caption{\textbf{Indoor and outdoor evaluation under different training-domain settings.}
Indoor combines HSSD and INS, whereas Outdoor combines iCity and Infinigen. Indoor, Outdoor, and Overall PSNR values are sample-weighted according to the number of test images.}
\label{tab:indoor_outdoor_evaluation}
\setlength{\tabcolsep}{2.6pt}
\resizebox{\columnwidth}{!}{%
\begin{tabular}{l l c c c}
\toprule
\multirow{2}{*}{Method}
& \multirow{2}{*}{Training Data}
& \multicolumn{1}{c}{Indoor}
& \multicolumn{1}{c}{Outdoor}
& \multicolumn{1}{c}{Overall} \\
\cmidrule(lr){3-3}
\cmidrule(lr){4-4}
\cmidrule(lr){5-5}
&
&
PSNR$\uparrow$
&
PSNR$\uparrow$
&
PSNR$\uparrow$ \\
\midrule
OmniSR
& INS (30K)
& 29.55
& 19.53
& 26.11 \\

OmniSR
& INS+iCity+Infinigen (36K)
& 29.47
& 26.53
& 28.46 \\
\midrule
Ours
& INS (30K)
& 30.46
& 18.82
& \textbf{26.47} \\

Ours
& INS+iCity+Infinigen (36K)
& 30.21
& 27.51
& \textbf{29.28}  \\
\bottomrule
\end{tabular}}
\end{table}


\subsection{Cross-Domain Generalization Analysis}
\label{sec:domain_analysis}

To analyze cross-domain generalization, we divide the test set into indoor scenes (INS and HSSD) and outdoor scenes (iCity and Infinigen). As shown in Table~\ref{tab:indoor_outdoor_evaluation}, we compare our method with OmniSR, which achieves the second-best performance on our WildShadow-I benchmark. When trained only on indoor data (INS), both methods perform well on indoor scenes but experience a significant performance drop on outdoor scenes, with PSNR values below 20~dB. This reveals a substantial domain gap between indoor and outdoor shadow removal, caused by differences in illumination conditions, shadow patterns, materials, and scene structures. These results indicate that models trained on limited indoor data struggle to generalize to diverse outdoor environments.

After extending the training data with outdoor scenes from iCity and Infinigen, both methods achieve substantial improvements on outdoor scenes. Specifically, our method improves the outdoor PSNR from 18.82 to 27.51~dB while maintaining comparable indoor performance (30.46 vs. 30.21~dB), resulting in an overall PSNR improvement from 26.47 to 29.28~dB. Moreover, our method consistently outperforms OmniSR under the multi-domain training setting, achieving higher indoor, outdoor, and overall performance. These results demonstrate that WildShadow-I effectively improves cross-domain generalization, and our framework can better leverage diverse scene distributions for robust shadow removal.

\begin{table}[t]
\centering
\caption{{Ablation of the proposed method.}}
\label{tab:ablation_design}
\small
\setlength{\tabcolsep}{5.0pt}
\begin{tabular}{l c c}
\toprule
Variant & PSNR$\uparrow$ & SSIM$\uparrow$ \\
\midrule
w/o depth condition & 29.10 & 0.959 \\
w/o detail injection module & 28.66 & 0.946 \\
w/o DA3 feature & 29.15 & 0.960 \\
w/o mask-aware freq. modulation & 29.16 & 0.960 \\
full & \textbf{29.28} & \textbf{0.961} \\
\bottomrule
\end{tabular}
\end{table}

\subsection{Ablation Study}
\label{sec:ablation}

We conduct component-wise ablations on the proposed WildShadow-I dataset under the same training and evaluation settings. Table~\ref{tab:ablation_design} evaluates the contributions of depth conditioning, the detail injection module, DA3 features, and mask-aware frequency modulation. Removing the depth condition decreases PSNR from 29.28 to 29.10~dB, demonstrating the effectiveness of geometric guidance for shadow removal. Removing the detail injection module causes the largest performance drop, with PSNR decreasing by 0.62~dB and SSIM by 0.015, indicating that latent diffusion alone is insufficient for fine-detail restoration.

Removing DA3 features and mask-aware frequency modulation leads to PSNR decreases of 0.13~dB and 0.12~dB, respectively. These results verify that geometry-aware guidance and frequency-aware modulation further improve detail recovery and suppress residual shadow artifacts. The full model achieves the best performance by combining all components.

\subsection{Real-World Evaluation}
\label{sec:real_world_evaluation}

We further evaluate our method on real-world scenes without paired shadow-free references. Figure~\ref{fig:dl3dv_compare} compares different methods on unseen real-world DL3DV videos. Since these scenes do not provide ground-truth shadow-free videos, we focus on residual shadows, illumination consistency, and structural preservation. OmniSR, StableSR, and PhaSR often leave noticeable dark regions or introduce illumination inconsistencies in challenging areas. The final results recover sharper boundaries and fine-grained details while preserving the corrected illumination. The enlarged regions further demonstrate the complementary effects of the video diffusion prior and the detail-restoration module on real captured scenes.

\subsection{Video Shadow Removal Evaluation}
\label{sec:video_evaluation}

\begin{table}[t]
\centering
\caption{Quantitative comparison on the video shadow removal benchmark. Image-based methods and frame-by-frame variants are evaluated independently on each frame. $E_{\mathrm{warp}}$ measures temporal warping error, with values reported in $\times10^{-3}$.}
\label{tab:video_frame_comparison}
\resizebox{\linewidth}{!}{
\begin{tabular}{lccc}
\toprule
Method & PSNR $\uparrow$ & SSIM $\uparrow$ & $E_{\mathrm{warp}}\downarrow$ \\
\midrule
OmniSR (frame-by-frame) 
& 29.809 
& 0.9536 
& 2.032 \\

PSTNet (video model) 
& 26.798
& 0.9403
& 1.998 \\

Ours (frame-by-frame) 
& 30.659 
& \textbf{0.9684} 
& 2.208 \\

Ours (video model) 
& \textbf{30.802} 
& 0.9655 
& \textbf{1.775} \\
\bottomrule
\end{tabular}}
\end{table}

We train a unified video diffusion model using the complete WildShadow video dataset together with the image dataset introduced in Table~\ref{tab:quantitative_comparison}, where each image is treated as a single-frame video. We evaluate our model on the video benchmark introduced in the dataset section. Following common practice in video restoration, PSNR and SSIM are averaged over frames within each clip and then across all clips. 

We compare our video model with a representative image-based method (OmniSR), a video shadow removal method (PSTNet), and our frame-by-frame variant. Besides reconstruction quality, we evaluate temporal stability using the warping error $E_{\mathrm{warp}}$~\cite{lai2018blind}, which measures the motion-compensated error between adjacent frames. As shown in Table~\ref{tab:video_frame_comparison}, our video model achieves competitive reconstruction quality while providing superior temporal consistency. In particular, it outperforms PSTNet in terms of PSNR, SSIM, and temporal warping error. Additional qualitative comparisons are provided in the supplementary video.

\section{Conclusion}
\label{sec:conclusion}

We present WildShadowRemover, a diffusion-based framework for in-the-wild video shadow removal, together with WildShadow, a large-scale synthetic video shadow removal dataset. By leveraging video diffusion priors, geometry-aware guidance, and shadow-mask-guided detail restoration, our method achieves effective shadow removal with improved detail preservation and temporal consistency.

\paragraph{Limitations.}
Our method still struggles with scenes containing highly complex geometries, such as dense foliage and thin structures, where intricate occlusions and fine-scale shadow patterns remain challenging. Moreover, our framework relies primarily on synthetic paired data for training, and how to effectively leverage large-scale real-world videos through self-supervised or weakly supervised learning remains an open problem.

\bibliography{main}

@String(CVPR= {IEEE Conf. Comput. Vis. Pattern Recog.})

@String(ECCV= {Eur. Conf. Comput. Vis.})

@String(TIP  = {IEEE Trans. Image Process.})

@String(ICLR = {Int. Conf. Learn. Represent.})

@String(AAAI = {AAAI})

@String(CVPR  = {CVPR})

@String(ECCV  = {ECCV})

@String(TIP   = {IEEE TIP})

@String(ICLR  = {ICLR})

@article{ling2023dl3dv,
  title={Dl3dv-10k: A large-scale scene dataset for deep learning-based 3d vision},
  author={Ling, Lu and Sheng, Yichen and Tu, Zhi and Zhao, Wentian and Xin, Cheng and Wan, Kun and Yu, Lantao and Guo, Qianyu and Yu, Zixun and Lu, Yawen and others},
  journal={arXiv preprint arXiv:2312.16256},
  volume={3},
  year={2023}
}

@inproceedings{hu2022lora,
  title={LoRA: Low-Rank Adaptation of Large Language Models},
  author={Hu, Edward J. and Shen, Yelong and Wallis, Phillip and Allen-Zhu, Zeyuan and Li, Yuanzhi and Wang, Shean and Wang, Lu and Chen, Weizhu},
  booktitle={International Conference on Learning Representations (ICLR)},
  year={2022}
}

@article{xu2025diffusion,
  title={Diffusion Knows Transparency: Repurposing Video Diffusion for Transparent Object Depth and Normal Estimation},
  author={Xu, Shaocong and Wei, Songlin and Wei, Qizhe and Geng, Zheng and Li, Hong and Shen, Licheng and Sun, Qianpu and Han, Shu and Ma, Bin and Li, Bohan and others},
  journal={arXiv preprint arXiv:2512.23705},
  year={2025}
}

@article{xiao2026relit,
  title={Relit-LiVE: Relight Video by Jointly Learning Environment Video},
  author={Xiao, Weiqing and Li, Hong and Yang, Xiuyu and Chen, Houyuan and Li, Wenyi and Liu, Tianqi and Xu, Shaocong and Ye, Chongjie and Zhao, Hao and Wang, Beibei},
  journal={arXiv preprint arXiv:2605.06658},
  year={2026}
}

@inproceedings{yu2025objectmover,
  title={Objectmover: Generative object movement with video prior},
  author={Yu, Xin and Wang, Tianyu and Kim, Soo Ye and Guerrero, Paul and Chen, Xi and Liu, Qing and Lin, Zhe and Qi, Xiaojuan},
  booktitle={Proceedings of the IEEE/CVF Conference on Computer Vision and Pattern Recognition},
  pages={17682--17691},
  year={2025}
}

@inproceedings{xu2025detail,
  title={Detail-preserving latent diffusion for stable shadow removal},
  author={Xu, Jiamin and Zheng, Yuxin and Li, Zelong and Wang, Chi and Gu, Renshu and Xu, Weiwei and Xu, Gang},
  booktitle={Proceedings of the IEEE/CVF Conference on Computer Vision and Pattern Recognition},
  pages={7592--7602},
  year={2025}
}

@article{ho2022video,
  title={Video diffusion models},
  author={Ho, Jonathan and Salimans, Tim and Gritsenko, Alexey and Chan, William and Norouzi, Mohammad and Fleet, David J},
  journal={Advances in neural information processing systems},
  volume={35},
  pages={8633--8646},
  year={2022}
}

@article{wan2025wan,
  title={Wan: Open and advanced large-scale video generative models},
  author={Wan, Team and Wang, Ang and Ai, Baole and Wen, Bin and Mao, Chaojie and Xie, Chen-Wei and Chen, Di and Yu, Feiwu and Zhao, Haiming and Yang, Jianxiao and others},
  journal={arXiv preprint arXiv:2503.20314},
  year={2025}
}

@article{lin2025depth,
  title={Depth anything 3: Recovering the visual space from any views},
  author={Lin, Haotong and Chen, Sili and Liew, Junhao and Chen, Donny Y and Li, Zhenyu and Shi, Guang and Feng, Jiashi and Kang, Bingyi},
  journal={arXiv preprint arXiv:2511.10647},
  year={2025}
}

@inproceedings{raistrick2023infinite,
  title={Infinite photorealistic worlds using procedural generation},
  author={Raistrick, Alexander and Lipson, Lahav and Ma, Zeyu and Mei, Lingjie and Wang, Mingzhe and Zuo, Yiming and Kayan, Karhan and Wen, Hongyu and Han, Beining and Wang, Yihan and others},
  booktitle={Proceedings of the IEEE/CVF conference on computer vision and pattern recognition},
  pages={12630--12641},
  year={2023}
}

@misc{icity2024,
  title        = {iCity: Procedural City Generator for Blender},
  author       = {Hothifa Smair},
  year         = {2024},
  howpublished = {\url{https://icity3d.com}},
  note         = {Blender add-on}
}

@article{khanna2023hssd,
  author={{Khanna*}, Mukul and {Mao*}, Yongsen and Jiang, Hanxiao and Haresh, Sanjay and Shacklett, Brennan and Batra, Dhruv and Clegg, Alexander and Undersander, Eric and Chang, Angel X. and Savva, Manolis},
  title={{Habitat Synthetic Scenes Dataset (HSSD-200): An Analysis of 3D Scene Scale and Realism Tradeoffs for ObjectGoal Navigation}},
  journal={arXiv preprint},
  year={2023},
  eprint={2306.11290},
  archivePrefix={arXiv},
  primaryClass={cs.CV}
}

@inproceedings{zhang2021noshadow,
  title={No Shadow Left Behind: Removing Objects and Their Shadows Using Approximate Lighting and Geometry},
  author={Zhang, Ling and others},
  booktitle={Proceedings of the IEEE/CVF Conference on Computer Vision and Pattern Recognition (CVPR)},
  year={2021}
}

@inproceedings{infinigen2024indoors,
    author    = {Raistrick, Alexander and Mei, Lingjie and Kayan, Karhan and Yan, David and Zuo, Yiming and Han, Beining and Wen, Hongyu and Parakh, Meenal and Alexandropoulos, Stamatis and Lipson, Lahav and Ma, Zeyu and Deng, Jia},
    title     = {Infinigen Indoors: Photorealistic Indoor Scenes using Procedural Generation},
    booktitle = {Proceedings of the IEEE/CVF Conference on Computer Vision and Pattern Recognition (CVPR)},
    month     = {June},
    year      = {2024},
    pages     = {21783-21794}
}

@inproceedings{hu2019mask,
  title={Mask-shadowgan: Learning to remove shadows from unpaired data},
  author={Hu, Xiaowei and Jiang, Yitong and Fu, Chi-Wing and Heng, Pheng-Ann},
  booktitle={Proceedings of the IEEE/CVF international conference on computer vision},
  pages={2472--2481},
  year={2019}
}

@inproceedings{wang2018stacked,
  title={Stacked conditional generative adversarial networks for jointly learning shadow detection and shadow removal},
  author={Wang, Jifeng and Li, Xiang and Hui,Le and Yang, Jian},
  booktitle={Proceedings of the IEEE conference on computer vision and pattern recognition},
  pages={1788--1797},
  year={2018}
}

@inproceedings{zhu2022bijective,
  title={Bijective mapping network for shadow removal},
  author={Zhu, Yurui and Huang, Jie and Fu, Xueyang and Zhao, Feng and Sun, Qibin and Zha, Zheng-Jun},
  booktitle={Proceedings of the IEEE/CVF Conference on Computer Vision and Pattern Recognition},
  pages={5627--5636},
  year={2022}
}

@inproceedings{fu2021auto,
  title={Auto-exposure fusion for single-image shadow removal},
  author={Fu, Lan and Zhou, Changqing and Guo, Qing and Juefei-Xu, Felix and Yu, Hongkai and Feng, Wei and Liu, Yang and Wang, Song},
  booktitle={Proceedings of the IEEE/CVF conference on computer vision and pattern recognition},
  pages={10571--10580},
  year={2021}
}

@article{guo2023shadowformer,
  title={Shadowformer: Global context helps image shadow removal},
  author={Guo, Lanqing and Huang, Siyu and Liu, Ding and Cheng, Hao and Wen, Bihan},
  journal={arXiv preprint arXiv:2302.01650},
  year={2023}
}

@article{liu2023decoupled,
  title={A Decoupled Multi-Task Network for Shadow Removal},
  author={Liu, Jiawei and Wang, Qiang and Fan, Huijie and Li, Wentao and Qu, Liangqiong and Tang, Yandong},
  journal={IEEE Transactions on Multimedia},
  year={2023},
  publisher={IEEE}
}

@inproceedings{cun2020towards,
  title={Towards ghost-free shadow removal via dual hierarchical aggregation network and shadow matting gan},
  author={Cun, Xiaodong and Pun, Chi-Man and Shi, Cheng},
  booktitle={Proceedings of the AAAI Conference on Artificial Intelligence},
  volume={34},
  number={07},
  pages={10680--10687},
  year={2020}
}

@inproceedings{zhu2022efficient,
  title={Efficient model-driven network for shadow removal},
  author={Zhu, Yurui and Xiao, Zeyu and Fang, Yanchi and Fu, Xueyang and Xiong, Zhiwei and Zha, Zheng-Jun},
  booktitle={Proceedings of the AAAI conference on artificial intelligence},
  volume={36},
  number={3},
  pages={3635--3643},
  year={2022}
}

@inproceedings{qu2017deshadownet,
  title={Deshadownet: A multi-context embedding deep network for shadow removal},
  author={Qu, Liangqiong and Tian, Jiandong and He, Shengfeng and Tang, Yandong and Lau, Rynson WH},
  booktitle={Proceedings of the IEEE Conference on Computer Vision and Pattern Recognition},
  pages={4067--4075},
  year={2017}
}

@inproceedings{le2020shadow,
  title={From shadow segmentation to shadow removal},
  author={Le, Hieu and Samaras, Dimitris},
  booktitle={Computer Vision--ECCV 2020: 16th European Conference, Glasgow, UK, August 23--28, 2020, Proceedings, Part XI 16},
  pages={264--281},
  year={2020},
  organization={Springer}
}

@inproceedings{guo2023shadowdiffusion,
  title={Shadowdiffusion: When degradation prior meets diffusion model for shadow removal},
  author={Guo, Lanqing and Wang, Chong and Yang, Wenhan and Huang, Siyu and Wang, Yufei and Pfister, Hanspeter and Wen, Bihan},
  booktitle={Proceedings of the IEEE/CVF Conference on Computer Vision and Pattern Recognition},
  pages={14049--14058},
  year={2023}
}

@article{wang2004image,
  title={Image quality assessment: from error visibility to structural similarity},
  author={Wang, Zhou and Bovik, Alan C and Sheikh, Hamid R and Simoncelli, Eero P},
  journal={IEEE TIP},
  volume={13},
  number={4},
  pages={600--612},
  year={2004},
}

@inproceedings{sanin2010improved,
  title={Improved shadow removal for robust person tracking in surveillance scenarios},
  author={Sanin, Andres and Sanderson, Conrad and Lovell, Brian C},
  booktitle={2010 20th International Conference on Pattern Recognition},
  pages={141--144},
  year={2010},
  organization={IEEE}
}

@inproceedings{li2018learning,
  title={Learning intrinsic image decomposition from watching the world},
  author={Li, Zhengqi and Snavely, Noah},
  booktitle={Proceedings of the IEEE conference on computer vision and pattern recognition},
  pages={9039--9048},
  year={2018}
}

@inproceedings{nestmeyer2020learning,
  title={Learning physics-guided face relighting under directional light},
  author={Nestmeyer, Thomas and Lalonde, Jean-Fran{\c{c}}ois and Matthews, Iain and Lehrmann, Andreas},
  booktitle={Proceedings of the IEEE/CVF Conference on Computer Vision and Pattern Recognition},
  pages={5124--5133},
  year={2020}
}

@inproceedings{ye2023intrinsicnerf,
  title={Intrinsicnerf: Learning intrinsic neural radiance fields for editable novel view synthesis},
  author={Ye, Weicai and Chen, Shuo and Bao, Chong and Bao, Hujun and Pollefeys, Marc and Cui, Zhaopeng and Zhang, Guofeng},
  booktitle={Proceedings of the IEEE/CVF International Conference on Computer Vision},
  pages={339--351},
  year={2023}
}

@inproceedings{le2019shadow,
  title={Shadow removal via shadow image decomposition},
  author={Le, Hieu and Samaras, Dimitris},
  booktitle={Proceedings of the IEEE/CVF International Conference on Computer Vision},
  pages={8578--8587},
  year={2019}
}

@inproceedings{fu20213dfront,
  title={3d-front: 3d furnished rooms with layouts and semantics},
  author={Fu, Huan and Cai, Bowen and Gao, Lin and Zhang, Ling-Xiao and Wang, Jiaming and Li, Cao and Zeng, Qixun and Sun, Chengyue and Jia, Rongfei and Zhao, Binqiang and others},
  booktitle={Proceedings of the IEEE/CVF International Conference on Computer Vision},
  pages={10933--10942},
  year={2021}
}

@inproceedings{li2022physically,
  title={Physically-based editing of indoor scene lighting from a single image},
  author={Li, Zhengqin and Shi, Jia and Bi, Sai and Zhu, Rui and Sunkavalli, Kalyan and Ha{\v{s}}an, Milo{\v{s}} and Xu, Zexiang and Ramamoorthi, Ravi and Chandraker, Manmohan},
  booktitle={European Conference on Computer Vision},
  pages={555--572},
  year={2022},
  organization={Springer}
}

@article{dong2024shadowrefiner,
  title={ShadowRefiner: Towards mask-free shadow removal via fast fourier transformer},
  author={Dong, Wei and Zhou, Han and Tian, Yuqiong and Sun, Jingke and Liu, Xiaohong and Zhai, Guangtao and Chen, Jun},
  journal={arXiv preprint arXiv:2406.02559},
  year={2024}
}

@article{zhang1999shape,
  title={Shape-from-shading: a survey},
  author={Zhang, Ruo and Tsai, Ping-Sing and Cryer, James Edwin and Shah, Mubarak},
  journal={IEEE transactions on pattern analysis and machine intelligence},
  volume={21},
  number={8},
  pages={690--706},
  year={1999},
  publisher={IEEE}
}

@inproceedings{luo2023refusion,
  title={Refusion: Enabling large-size realistic image restoration with latent-space diffusion models},
  author={Luo, Ziwei and Gustafsson, Fredrik K and Zhao, Zheng and Sj{\"o}lund, Jens and Sch{\"o}n, Thomas B},
  booktitle={Proceedings of the IEEE/CVF conference on computer vision and pattern recognition},
  pages={1680--1691},
  year={2023}
}

@article{luo2024diff,
  title={Diff-Shadow: Global-guided Diffusion Model for Shadow Removal},
  author={Luo, Jinting and Li, Ru and Jiang, Chengzhi and Han, Mingyan and Zhang, Xiaoming and Jiang, Ting and Fan, Haoqiang and Liu, Shuaicheng},
  journal={arXiv preprint arXiv:2407.16214},
  year={2024}
}

@inproceedings{mei2024latent,
  title={Latent feature-guided diffusion models for shadow removal},
  author={Mei, Kangfu and Figueroa, Luis and Lin, Zhe and Ding, Zhihong and Cohen, Scott and Patel, Vishal M},
  booktitle={Proceedings of the IEEE/CVF Winter Conference on Applications of Computer Vision},
  pages={4313--4322},
  year={2024}
}

@inproceedings{xiao2024homoformer,
  title={HomoFormer: Homogenized Transformer for Image Shadow Removal},
  author={Xiao, Jie and Fu, Xueyang and Zhu, Yurui and Li, Dong and Huang, Jie and Zhu, Kai and Zha, Zheng-Jun},
  booktitle={Proceedings of the IEEE/CVF Conference on Computer Vision and Pattern Recognition},
  pages={25617--25626},
  year={2024}
}

@inproceedings{jin2024des3,
  title={DeS3: Adaptive Attention-Driven Self and Soft Shadow Removal Using ViT Similarity},
  author={Jin, Yeying and Ye, Wei and Yang, Wenhan and Yuan, Yuan and Tan, Robby T},
  booktitle={Proceedings of the AAAI Conference on Artificial Intelligence},
  volume={38},
  number={3},
  pages={2634--2642},
  year={2024}
}

@inproceedings{zhang2017video,
  title={Video Shadow Removal Using Spatio-temporal Illumination Transfer},
  author={Zhang, J. and others},
  booktitle={Proceedings of the IEEE International Conference on Computer Vision Workshops (ICCVW)},
  year={2017}
}

@article{chen2024pstnet,
  title={Learning Physical-Spatio-Temporal Features for Video Shadow Removal},
  author={Chen, Zhihao and Wan, Liang and Xiao, Yefan and Zhu, Lei and Fu, Huazhu},
  journal={IEEE Transactions on Circuits and Systems for Video Technology},
  volume={34},
  number={7},
  pages={5830--5842},
  year={2024},
  publisher={IEEE}
}

@inproceedings{lai2018blind,
  title={Learning Blind Video Temporal Consistency},
  author={Lai, Wei-Sheng and Huang, Jia-Bin and Wang, Oliver and Shechtman, Eli and Yang, Ming-Hsuan},
  booktitle={ECCV},
  year={2018}
}

@inproceedings{xu2024omnisr,
  title={Omnisr: Shadow removal under direct and indirect lighting},
  author={Xu, Jiamin and Li, Zelong and Zheng, Yuxin and Huang, Chenyu and Gu, Renshu and Xu, Weiwei and Xu, Gang},
  booktitle={Proceedings of the AAAI Conference on Artificial Intelligence},
  volume={39},
  number={8},
  pages={8887--8895},
  year={2025}
}

@inproceedings{wang2018esrgan,
  title={Esrgan: Enhanced super-resolution generative adversarial networks},
  author={Wang, Xintao and Yu, Ke and Wu, Shixiang and Gu, Jinjin and Liu, Yihao and Dong, Chao and Qiao, Yu and Loy,Chen Change},
  booktitle={Proceedings of the European conference on computer vision (ECCV) workshops},
  pages={63-–79},
  year={2018}
}

@inproceedings{phaser2026,
  author    = {Lee, Chia-Ming and Lin, Yu-Fan and Hsiao, Yu-Jou and Jiang, Jin-Hui and Liu, Yu-Lun and Hsu, Chih-Chung},
  title     = {PhaSR: Generalized Image Shadow Removal with Physically Aligned Priors},
  booktitle = {Proceedings of the IEEE/CVF Conference on Computer Vision and Pattern Recognition (CVPR)},
  pages     = {22679--22688},
  year      = {2026}
}

@inproceedings{densesr2025,
  author    = {Lin, Yu-Fan and Lee, Chia-Ming and Hsu, Chih-Chung},
  title     = {DenseSR: Image Shadow Removal as Dense Prediction},
  booktitle = {Proceedings of the 33rd ACM International Conference on Multimedia},
  year      = {2025}
}

\end{document}